\documentclass[letterpaper, 10 pt, conference]{ieeeconf}  % Comment this line out if you need a4paper

\IEEEoverridecommandlockouts                              % This command is only needed if 
\usepackage{amsfonts}
\let\labelindent\relax
\usepackage{enumitem}
\usepackage{booktabs}
\usepackage{amsmath}
\usepackage{algorithm} 
\usepackage{graphicx}
\usepackage{subfig}
\usepackage[margin=0.792in]{geometry}
\usepackage{caption}
\usepackage{bbm}
\usepackage{algpseudocode} 
\usepackage{multirow}
\usepackage{xcolor}
\algnewcommand{\Inputs}[1]{%
  \State \textbf{Inputs:}
  \Statex \hspace*{\algorithmicindent}\parbox[t]{.8\linewidth}{\raggedright #1}
}
\algnewcommand{\Initialize}[1]{%
  \State \textbf{Initialize:}
  \Statex \hspace*{\algorithmicindent}\parbox[t]{.8\linewidth}{\raggedright #1}
}
\newcommand\Tstrut{\rule{0pt}{2.6ex}}         % = `top' strut
\newcommand\Bstrut{\rule[-0.9ex]{0pt}{0pt}}   % = `bottom' strut

\DeclareMathOperator*{\argmin}{\arg\!\min}
\DeclareMathOperator*{\argmax}{\arg\!\max}
\usepackage{cite}
\usepackage{amsmath,amssymb,amsfonts}
\usepackage{graphicx}
\usepackage{textcomp}
\usepackage{xcolor}
\def\BibTeX{{\rm B\kern-.05em{\sc i\kern-.025em b}\kern-.08em
    T\kern-.1667em\lower.7ex\hbox{E}\kern-.125emX}}
\title{\LARGE \bf
Neural Rearrangement Planning for Object Retrieval from Confined Spaces Perceivable by Robot's In-hand RGB-D Sensor
}

\author{Hanwen Ren and Ahmed H. Qureshi% <-this % stops a space
\thanks{*This work was supported by the National Science Foundation (NSF) under award no. 2204528.}
\thanks{Hanwen Ren and Ahmed H. Qureshi are with the Department of Computer Science, Purdue University, West Lafayette, IN, USA, 47907. Email: {\tt\small$\{$ren221, ahqureshi$\}@$purdue.edu}}% <-this % stops a space
}

\begin{document}

\maketitle
\thispagestyle{empty}
\pagestyle{empty}

%%%%%%%%%%%%%%%%%%%%%%%%%%%%%%%%%%%%%%%%%%%%%%%%%%%%%%%%%%%%%%%%%%%%%%%%%%%%%%%%
\begin{abstract}

Rearrangement planning for object retrieval tasks from confined spaces is a challenging problem, primarily due to the lack of open space for robot motion and limited perception. Several traditional methods exist to solve object retrieval tasks, but they require overhead cameras for perception and a time-consuming exhaustive search to find a solution and often make unrealistic assumptions, such as having identical, simple geometry objects in the environment. This paper presents a neural object retrieval framework that efficiently performs rearrangement planning of unknown, arbitrary objects in confined spaces to retrieve the desired object using a given robot grasp. Our method actively senses the environment with the robot's in-hand camera. It then selects and relocates the non-target objects such that they do not block the robot path homotopy to the target object, thus also aiding an underlying path planner in quickly finding robot motion sequences. Furthermore, we demonstrate our framework in challenging scenarios, including real-world cabinet-like environments with arbitrary household objects. The results show that our framework achieves the best performance among all presented methods and is, on average, two orders of magnitude computationally faster than the best-performing baselines.

\end{abstract}

\section{Introduction}
Target object retrieval from unknown confined spaces is critical for robots intending to assist people in their daily lives \cite{yamazaki2012home}. For instance, robots aiding at hospitals will often have to retrieve the required medicines from cabinets. At factory floors, this could involve fetching tools from toolboxes. Additionally, these robot skills are also desirable for search and rescue at disaster sites to help the affected people \cite{schlenoff2005robot}. However, object retrieval from unknown confined spaces imposes significant challenges. First, scene observation is more complex in these settings than in traditional tabletop environments due to the heavy occlusion between objects, imperfect light conditions, and limited camera view angles. Second, the robot needs to clear the pathway to the target by relocating other objects in the scene. Third, robot grasping also imposes challenges as oftentimes only limited stable grasp candidates are available.
\begin{figure}[ht]
    \centering
    \includegraphics[trim = {0cm 0cm 0cm 0cm}, clip, width = 8.5cm]{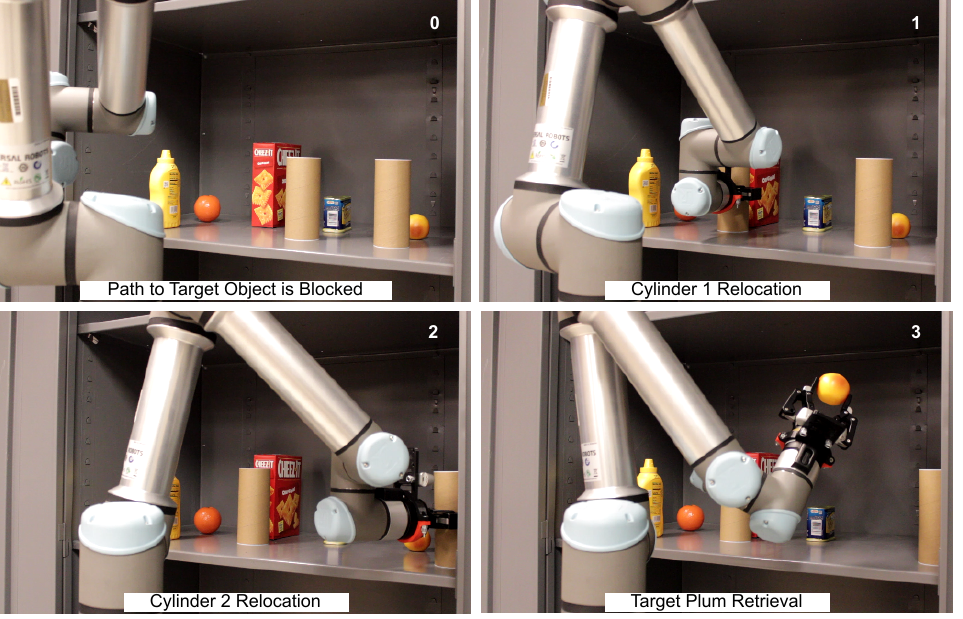}
    \caption{Execution for retrieving the target object (``plum''): The robot's pathway to the target object is blocked in the initial setup. After executing the object manipulation plan from our method of relocating the pathway-blocking objects, in this case, the cylinders, the robot arm finally retrieves the plum from the confined cabinet environment.}
    \label{fig:real-experiment2}
    \vspace{-0.1in}
\end{figure}
One of the challenging components of rearrangement planning for object retrieval, besides its NP-hard complexity \cite{wilfong1988motion}, is that the algorithm needs to explicitly select path-blocking objects from a given observation and relocate them to  a feasible region so that the target can be accessed \cite{ahn2021integrated}. The existing approaches in task and motion planning (TAMP) \cite{garrett2021integrated} mainly try to solve this problem using tree-search-based methods \cite{ren2021extended} until the path to the target is obtained. However, these methods are computationally expensive; therefore, they are inapplicable for real-time applications requiring fast solutions \cite{zhou2022review}. In addition, current approaches \cite{vieira2022persistent} tend to model the problem in the 2D plane and assume the scene is fully observed in advance.

Inspired by recent developments of deep learning and their application in various rearrangement planning problems \cite{qureshi2021nerp, goyal2022ifor}, this paper presents a neural rearrangement planning approach for object retrieval from unknown confined spaces. Our system gathers observations of the unknown, confined environment with an in-hand camera via the active sensing approach \cite{10101696}. Once the target object is detected and is unretrievable, our framework switches to the rearrangement phase for making the pathway to reach the target object. During rearrangement, our method iteratively selects and relocates non-target objects within a confined space until the target object is kinematically reachable for the robot arm without collision. Using an in-hand camera makes it even more challenging to access confined spaces without collisions. In summary, the main contribution and salient features of our approach include the following:
%{\color{red} Our system gathers observations of the unknown, confined environment via the active sensing approach \cite{10101696}. Suppose the target object is detected but cannot be directly retrieved. In that case, our novel multimodal Transformer-based object selection neural network first picks one non-target object with the most significant chance of blocking the robot's pathway to the target. The selected object is then rearranged from the target to a kinematically feasible empty place proposed by our novel region proposal framework consisting of multimodal Transformer-based encoder-decoder neural networks.} The procedure is repeated until the target object is successfully retrieved. The main contribution and salient features of our approach are summarized as follows:
\begin{itemize}
\item A novel object selection framework that learns the nature of robot movement in confined environments and chooses the object that has the highest chance of blocking the robot's way of reaching the desired target.
\item A novel region proposal framework that is aware of robot path homotopy to the target object and, therefore, selects the relocation region for the given object such it will no longer block robot pathways.
\item A new data generation strategy capturing the robot path homotopy to given targets for training the object selection and region proposal network.
\item A unified framework for efficiently solving object retrieval tasks from unknown confined spaces with demonstrations in complex real-world cabinet-like scenarios. 
\end{itemize}
%To the best of our knowledge, this paper presents the first approach to solving object retrieval tasks in unknown confined spaces under object grasping constraints limiting the robot's motion. We evaluate our approach in various complex settings and compare them with several baseline methods. The result shows that our approach exhibits significantly low computation times in solving the object retrieval task and generalizes to real-world scenarios.
\setlength{\textfloatsep}{0pt}
\begin{figure*}[ht]
    \centering
    \includegraphics[trim = {2.5cm 5.2cm 0cm 22cm}, clip, width = 18.5cm]{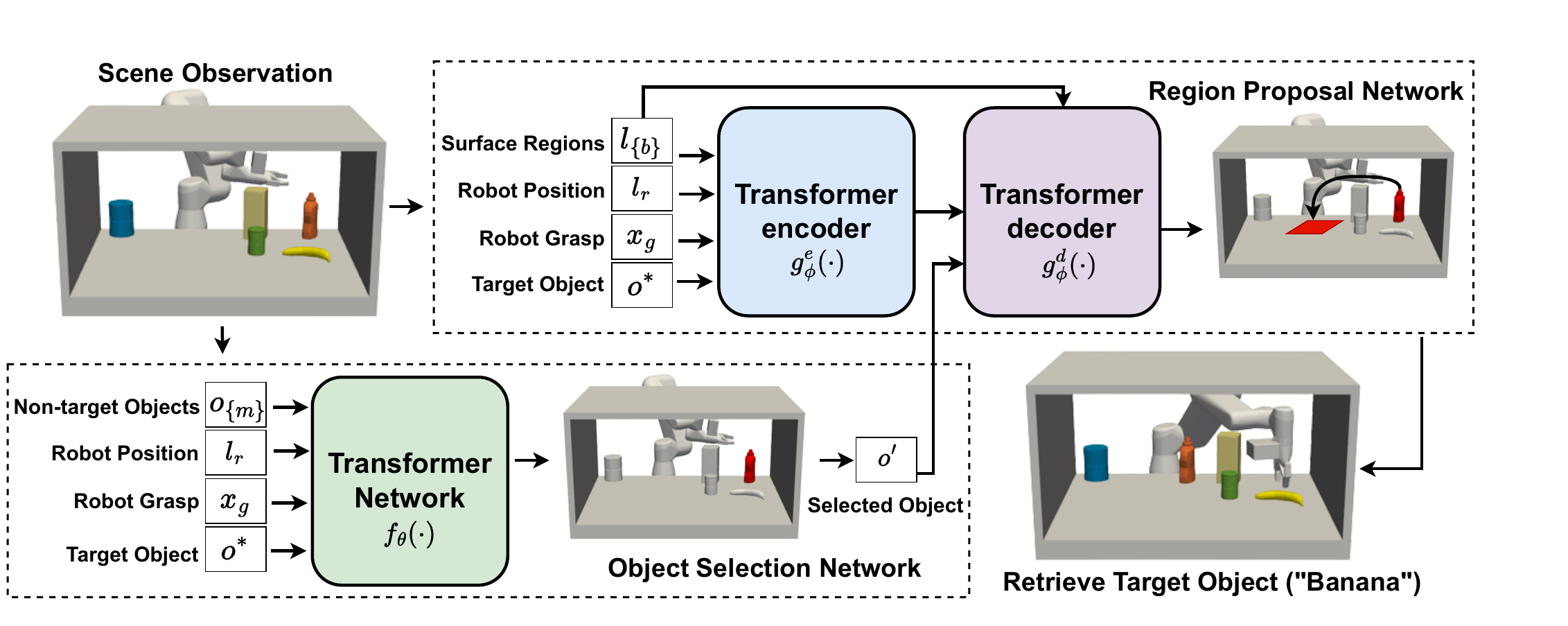}
    \caption{Neural Object Retrieval: The task is to retrieve the yellow object (Banana). Our main modules include the object selection and region proposal network. We do not show MLPs that encode the given inputs for brevity. Given the scene observation via active sensing, the object selection network selects the non-target object for rearrangement. The chosen object $o'$ is indicated in red in the bottom scene image. The region proposal network proposes the best placement region for the selected object to clear the pathway for target object retrieval. The proposed placement region on the environment surface is marked red on the top right scene image. Our robot moves the selected object to its new placement and retrieves the target object if possible otherwise repeats the object rearrangement process in the confined environment.}
    \label{fig:pipeline-figure}
\vspace{-0.2in}
\end{figure*}
%\vspace{-0.5in}

\section{Related Work}
The first kind of object retrieval task is performed in cluttered tabletop, open scenes \cite{ren2022rearrangement, huang2021visual, xiao2019online}. For instance, \cite{ren2022rearrangement} presents an object searching and retrieving system that leverages dynamically controlled sampling-based algorithms and extends to different robot tasks such as grasping, relocating, and sorting. Another approach \cite{nguyen2020robot} brings up a system that retrieves the desired object based on natural language instructions. However, the open space and fully observed environment settings make it much easier for the robot to fulfill the objective successfully.%Most of these works focus more on other aspects rather than explicit object retrieval via rearrangement planning.
\par
Many existing approaches \cite{ahn2021integrated, cheong2020relocate} also try to perform object retrieval in confined spaces. For instance, \cite{vieira2022persistent} uses non-prehensile actions such as pushing to clear up the space and grasp the target objects in cluttered environments. By utilizing non-prehensile grasp actions, the robot can interact with multiple objects simultaneously while eliminating the heavy computation involved in prehensile, pick-place, operations. Another approach \cite{lee2019efficient} aims to minimize the number of objects to relocate by choosing modified Vector Field Histogram-plus (VFH+) \cite{sary2018design} as a local planner to relocate non-target object at each step until a collision-free path to the target is obtained. In \cite{wang2022lazy}, the authors bring up a lazy object rearrangement planner that bypasses the extensive motion planning and collision-checking queries but only checks collision when a solution is found. However, all the methods mentioned above plan the object retrieval task in the fully observed 2D space consisting of identical objects.
%Then the only restriction the confined space brings is to limiting the robot's movement in a limited reaching angle in a 2D plane. 
%Furthermore, their assumptions of knowing complete environment information in a constrained and confined environment before robot motion planning is less likely to happen in real-world scenarios. %In addition, in their problem definition, all objects are uniformly-sized cylinders with locations known to the agent before planning. 
\par
Aside from directly solving the object retrieval tasks, other relevant works treat it as an instance in the rearrangement planning problem. For instance, several studies have been conducted in object rearrangement planning tasks with specific goal configurations in tabletop environments using tree-search \cite{hart1968formal, russell2010artificial} and randomized algorithms \cite{mirabel2016hpp, stilman2007manipulation}. Similarly, \cite{labbe2020monte} solves this problem by utilizing MCTS to avoid the expensive computation involved in traditional search-based methods. Works like \cite{qureshi2021nerp, goyal2022ifor, zeng2020transporter} use deep neural networks to tackle different sub-tasks engaged in the process. %In NePR, graph neural networks provide scene understanding and pair the same object instance at the start/goal configuration. At each step, object selection and region proposal are also generated by different MLPs. 
Some of those methods also generalize to real-world scenarios with never-before-seen objects. %Their approaches are trained in the simulation environment only but have the generalization capability to real-world scenes with never-seen-before Objects. 
However, these methods, due to the tabletop environment settings, cannot be deployed directly in confined spaces where object reachability, grasping, and robot motion planning induce several challenges for finding rearrangement solutions for object retrieval.

\section{Proposed Method}
This section formally presents our Neural Object Retrieval framework comprising the objective function, neural models, algorithm pipeline, and other implementation details. The framework structure is shown in Fig. \ref{fig:pipeline-figure}.
\subsection{Problem Definition}
Let a given scene be denoted as $S$ whose outline dimensions $d_x$, $d_y$, and $d_z$ along $x$, $y$, and $z$ axes, respectively, are assumed to be known. The scene $S$ comprises the observed space denoted as $S_o$ and an unobserved space denoted as $S_{ou} = S\backslash S_o$. A robot manipulator system situated at location $l_r \in \mathbb{R}^3$ with an RGB-D camera attached to its end-effector actively senses the given scene until the scene is fully observed, i.e., $S_o\approx S$. In the remainder of the paper, we use the notation $A_{\{B\}}$ to represent any arbitrary list $A$ containing $B \in \mathbb{N}$ number of elements. Once the scene is observed, we extract the observed and unobserved regions in the environment, denoted as $l_{\{b\}}=\{l_0,l_1,\cdots,l_b\}$. Each region is $n\times n$ grid on the given environment's ground surface. The $l_i \in l_{\{b\}}$ is a 4D vector,$(x,y,z,\mathrm{flag})$, with $x$,$y$, and $z$ corresponding to the spatial center position of the region and the $\mathrm{flag}$ indicating if that region was observed ($\mathrm{flag}=1$) or unobserved ($\mathrm{flag}=0$). The robot system configuration space is denoted as $\mathcal{Q}$ with obstacle and obstacle-free space indicated as $\mathcal{Q}_{free}$ and $\mathcal{Q}_{obs}=\mathcal{Q}\backslash \mathcal{Q}_{free}$, respectively. Given a collision-free robot configuration, $q \in \mathcal{Q}_{free}$, the corresponding gripper pose is indicated as $x_g \in SE(3)$. \par
%and in-hand camera poses are denoted as $x_q \in SE(3)$ and $x_c \in SE(3)$, respectively.%, each of six dimensions containing translation and orientation. 
%The active sensing module selects the sequence of camera viewpoints $\{x^0_c, x^1_c, \cdot, x^N_c\}$ to construct the fully-observed scene, i.e., $S_o \approx S$. 
The observed scene contains $m+1 \in \mathbb{N}$ objects, comprising non-target objects $o_{\{m\}} = \{o_1, o_2,...,o_m\}$ and a target object $o^*$ that is to be retrieved by the robot system. The scene state $S_o(t)$, at time $t$, indicates the state of objects $o_{\{m+1\}}$. The state of each object $o$ is a 6D vector, i.e., $(cx_o, cy_o, cz_o, dx_o, dy_o, dz_o)$, where the first three values capture the geometric center while the remainder denotes the dimension of its tightest axis-aligned bounding box. Let an indicator function $I(o^*, x^*_g, S_o(t))$ output 1 when the target object, $o^*$, is kinematically reachable without collision using the given grasp pose $x^*_g$; otherwise, 0 for the given scene state $S_o(t)$ at time $t$. 

We assume environment settings that are confined and cluttered and require a robot to rearrange non-target objects $o_{\{m\}}$ for reaching and retrieving the target object $o^*$ without collision. Therefore, a policy $\pi$ takes the scene observation $S_o(t)$ and robot state, at time $t$, selects an object $o'_i \in o_{\{m\}}$, whose current location is $l_{o'_i}^s \in \mathbb{R}^3$, and proposes an alternative placement $l^g_{o'_i} \in l_{\{b\}}$ for the selected object to clear a way to retrieve the target object $o^*$. Let function $d(l_r,l^s_{o'_i},l^g_{o'_i})$ represent the total Euclidean distance from robot base position $l_r$ to selected object location $l^s_{o'_i}$ and further to the placement region $l^g_{o'_i}$. The objective of our proposed work is to find an optimal policy $\pi^*$ that finds the shortest rearrangement sequence $\tau=\{(o'_1, l_{o'_1}^s, l_{o'_1}^g), (o'_2, l_{o'_2}^s, l_{o'_2}^g),...,(o'_T, l_{o'_T}^s, l_{o'_T}^g)\}$ to retrieve the target object $o^*$ using the given robot gripper pose $x^*_g$, i.e.,
\begin{equation}
\pi^*=\arg \max_\pi \mathbb{E}_{\tau\sim\pi}\bigg[\sum^T_{t=0}(I(o^*,x^*_g, S_o(t))-d(l_r,l^s_{o'_t},l^g_{o'_t}))\bigg]
\end{equation}
Therefore once the above objective function is optimized, an optimal policy will enable the target object retrieval by rearranging non-target objects with a minimum total move distance $d$ in the confined environments.

\subsection{Scene observation}
Since it is difficult to place an overhead camera in confined spaces to obtain observation, we utilize an active sensing (AS) approach \cite{10101696} that uses a robot with an in-hand camera to generate the scene perception. The scene is initially unknown except for its dimensions $(d_x, d_y, d_z)$. The AS method selects the best camera viewpoint sequences and registers all views into a unified scene observation. We run the AS module until the scene is fully observed and extract all the observed objects $o_{\{m+1\}} = \{o_1, o_2,...,o_m, o^*\}$ using scene segmentation \cite{wu2019detectron2}. Each object is represented with a 6D vector comprising its geometric center and bounding box dimensions. Such 6D object representation is more compact and computationally efficient than directly processing raw object point clouds using methods like PointNet++ \cite{qi2017pointnet++}. Finally, we also utilize the scene dimensions to determine the observed and unobserved regions $l_{\{b\}} = \{l_1, l_2,...,l_b\}$ in the given scene.
\subsection{Neural Object Selection}
We aim for a function that at each time step $t$ takes the scene observation $S_o(t)$, comprising objects $o_{\{m+1\}}$, and selects an object, $o' \in o_{\{m\}}$, that blocks the way for a homotopy of paths towards the target object, $o^*$, from the robot's current state. 
Directly running sampling-based motion planners (SMPs) \cite{elbanhawi2014sampling} can be time-consuming due to the low sampling efficiency caused by the strict collision constraints. Therefore,
checking for the path homotopy is crucial as it allows flexibility in choosing the underlying motion planner and increases its chances of successfully finding a path solution toward the given target.

Thus, we approach this problem by designing a neural network to learn the underlying nature of the homotopy paths and help the system quickly figure out the path-blocking objects.
%The neural network we proposed is called SelNet, which can be viewed as a function $f^s_{\theta}$ with parameters $\theta$ that outputs a predicted categorical distribution $\hat{p}_{\{m\}}$ representation the probability of each object being the optimal one based on the current scene information consisting of robot location $l_r$, gripper pose $q_g$, target object $o^*$ and all other observed objects $o_{\{m\}}$, 
The neural network we proposed is called Object Selection Network (OSNet), which can be viewed as a function $f_{\theta}$, with parameters $\theta$, that outputs a categorical probability distribution $\hat{p}_{\{m\}}$ over the non-target objects, representing the probability of each object blocking the way for path homotopy. The input to the OSNet is the current scene information consisting of objects state $(o_{\{m\}},o^*)$, robot location $l_r$, and desired gripper pose $x_g$ to grasp the target object, i.e.,
\begin{equation}
    \hat{p}_{\{m\}} \leftarrow f_{\theta}(\alpha_\theta(l_r), \gamma_\theta(x_g), \alpha_\theta (o^*), \alpha_\theta(o_{\{m\}}))
\end{equation}
where $m$ is the number of non-target objects, $l_r \in \mathbb{R}^3$, and $x_g \in SE(3)$. %Each object, including the target, is presented as a 6D vector comprising its geometric center and bounding box dimensions.  %, and  Compared with directly using the raw point clouds, this lightweight design helps the neural network to capture the most important object features and further reduce the time till network convergence. 
To process the given scene information, we use two different multi-layer perceptrons (MLP) \cite{haykin1994neural}, $\alpha_\theta$, $\gamma_\theta$ to embed robot $l_r$ plus objects $o_{\{m+1\}}$ and the gripper $x_g$, resulting in $m+3$ latent embeddings.
%To process the given scene information, we pass each of three inputs, $l_r$, $p_g$, and $o_{\{m+1\}}$, through different multi-layer perceptrons (MLP) \cite{haykin1994neural}, $\alpha_\theta$, $\gamma_\theta$, and $\kappa_\theta$, to generate $m+3$ latent embeddings. The MLP, $\kappa_\theta$, processing objects take each object state and output its embedding, resulting in their $m+1$ latent representations. 
All MLP latent embeddings form the input tokens for the multi-headed, self-attention Transformer network \cite{vaswani2017attention}, which captures the relationship between each embedding through its self-attention mechanism and outputs new $m+3$ representations. We extract the non-target $m$ objects' embeddings from the new representations and pass them through another MLP to predict the probability, $\hat{p} \in [0,1]$, of selecting each object for rearrangement. In summary, the function $f$ comprises two MLPs for encoding inputs, a Transformer Network for capturing input embedding relationships through self-attention, and a decoder MLP for predicting the selection probabilities for non-target objects. 
%\begin{algorithm}
%	\caption{Neural Object Retrieval Approach} 
%	\label{alg: neural object retrieval approach}
%	\begin{algorithmic}[1]
%         \Inputs{
%	    $l_r $ \Comment{robot location}\\
%         $x_g $ \Comment{target object grasp pose}\\
%         $o_{\{m\}} $ \Comment{Observed non-target objects state}\\
%        $o^* $ \Comment{Observed target object state}\\
%         $l_{\{b\}}$ \Comment{Surface regions}
%         }\vspace{0.1in}
%        
%	\While {not $I(o^*,x_g, l_r)$} \Comment{target object unreachable }
%                \State $o' \leftarrow \argmax_{o_{\{m\}}}\text{OSNet}(t_r, x_g, o^*, o_{\{m\}})$         %\Comment{Select object}
%                \State $l^g_{o'} \leftarrow \argmin_{l_{\{b\}}}\text{RPNet}(t_r, x_g, o^*, o', l_{\{b\}})$ 
%                \If {ComputePath$(l^s_{o'}, l^g_{o'})$}
%                        \State $\text{Rearrange}(o', l^g_{o'})$ \Comment{relocate the selected object {\color{red} if possible}} 
%                \EndIf 
%            \EndWhile\\
%            Retrieve target object $o^*$
%	\end{algorithmic} 
%\end{algorithm}

We train function $f_\theta$ end-to-end in a supervised manner using the Mean Squared Error (MSE) between the predicted, $\hat{p}_{\{m\}}$, and ground truth, $p_{\{m\}}$, probabilities. The ground truth, $p_{\{m\}}$, is a categorical distribution that is generated by leveraging the RRTConnect motion planner \cite{kuffner2000rrt} to select the object blocking the path homotopy. We run RRTConnect followed by path smoothing to compute multiple path to the target $o^*$ without considering collisions with other objects. Each non-target object receives a count whenever it blocks a path and the object with the maximum blocking counts should be the one to be relocated. Finally, the ground truth categorical distribution $p_{\{m\}}$ is formed by normalizing all the objects' blocking counts. %SelNet is trained in a supervised manner, and it aims to minimize the Mean Squared Error (MSE) between the prediction $\hat{p}_{\{m\}}$ and the ground truth $p_{\{m\}}$. 
During execution, we greedily select the object, $o' \in o_{\{m\}}$, with the maximum predicted probability in $\hat{p}_{\{m\}}$ for rearrangement.
%All latent embedding is concatenated into the initial latent scene representation $z'_s$, capturing the nature of the current environment. Next, a transformer encoder-based \cite{vaswani2017attention} neural network takes $z'_s$ and utilizes the multi-head self-attention mechanism to capture the relationship between each element. Each head uses learnable weight matrices $W^Q, W^K, W^V$ to project the input into different subspaces where the self-attention matrix can be calculated. This helps the network to reason between objects and their selection probability from different perspectives. The final scene representation $z_s$ is generated by passing $z'_s$ through several encoder blocks. After extracting the final object candidate embedding from $z_s$ based on its original indices in $z'_s$, we can get $\hat{p}_{\{m\}}$ by feeding them into a shared MLP. 

\subsection{Neural Rearrangement Region Proposal}
After selecting the object $o' \in o_{\{m\}}$ to rearrange, the next step is to decide its placement region, $l^g_{o'} \in l_{\{b\}}$, in the confined environment. The placement region for the selected object should be the nearest observed, kinematically reachable place that does not block the path homotopy towards the target. To achieve the above objectives, we propose a Region Proposal Network (RPNet), a neural function, $g_\phi$, with an encoder-decoder structure, comprising parameters $\phi$, that efficiently selects a feasible region for the placement of the selected object. Our architecture comprises three different MLPs, $\alpha_\phi, \gamma_\phi,$ and $\kappa_\phi$, to embed the robot location $l_r$, the target $(o^*)$ and selected object $(o')$ states, the desired gripper pose $x_g$, and the placement regions $l_{\{b\}}$, respectively. These MLPs are shared among our encoder, $g^e_\phi$, and decoder, $g^d_\phi$, modules. The encoder network is a multi-headed, self-attention Transformer Network that takes the MLP embedding of the robot's current location, target object, and placement regions, and outputs the latent representations $Z^e$, i.e.,
\begin{equation}
    Z^e \leftarrow  g^e_{\phi}\big(\alpha_\phi(l_r), \gamma_\phi(x_g), \alpha_\phi(o^*), \kappa_\phi (l_{\{b\}})\big)
\end{equation}

Therefore, the encoder encapsulates the object retrieval problem and its relation to all placement regions via self-attention. The decoder network is also a multi-headed, self-attention Transformer network, that takes the encoder network output, $Z^e$, and latent queries. The latent queries are formed by obtaining the MLP embeddings of the placement regions, $\kappa_\phi(l_{\{b\}})$, and adding each embedding with the MLP embedding of the selected object $\alpha_\phi(o')$. We assume the addition in latent space will enable our framework to foresee the impact of placing object $o'$ at a certain placement region $l \in l_{\{b\}}$. The latent output representation of all regions from the decoder network is sent to a MLP to obtain the final cost values, $\hat{c}_{\{b\}}$, indicating how well placement region and selected object pairs will aid in the target object retrieval, i.e., 
\begin{equation}
    \hat{c}_{\{b\}} \leftarrow g^d_{\phi}\big([\alpha_\phi(o')+\kappa_\phi(l_0)], \cdots,[\alpha_\phi(o') + \kappa_\phi(l_b)], Z^e\big)
\end{equation}

We train RPNet in a supervised manner. During the training phase, the network's objective is to minimize the MSE loss between the ground truth costs $c_{\{b\}}$ and the predicted $\hat{c}_{\{b\}}$ for a given object $o'$. Following the above-mentioned criteria, the occupied, unobserved, and robot path homotopy blocking regions get the maximum cost. For others, their cost values are assigned based on the Euclidean distance to the selected object's initial location. Once $g_\phi$ is trained using the ground truth labels for the variety of scenarios, we use it to select the placement region with the minimum cost for placing the selected object $o'$.

\begin{figure*}[ht]
    \centering
    \includegraphics[trim = {1cm 0.8cm 0cm 0cm}, clip, width = 18cm]{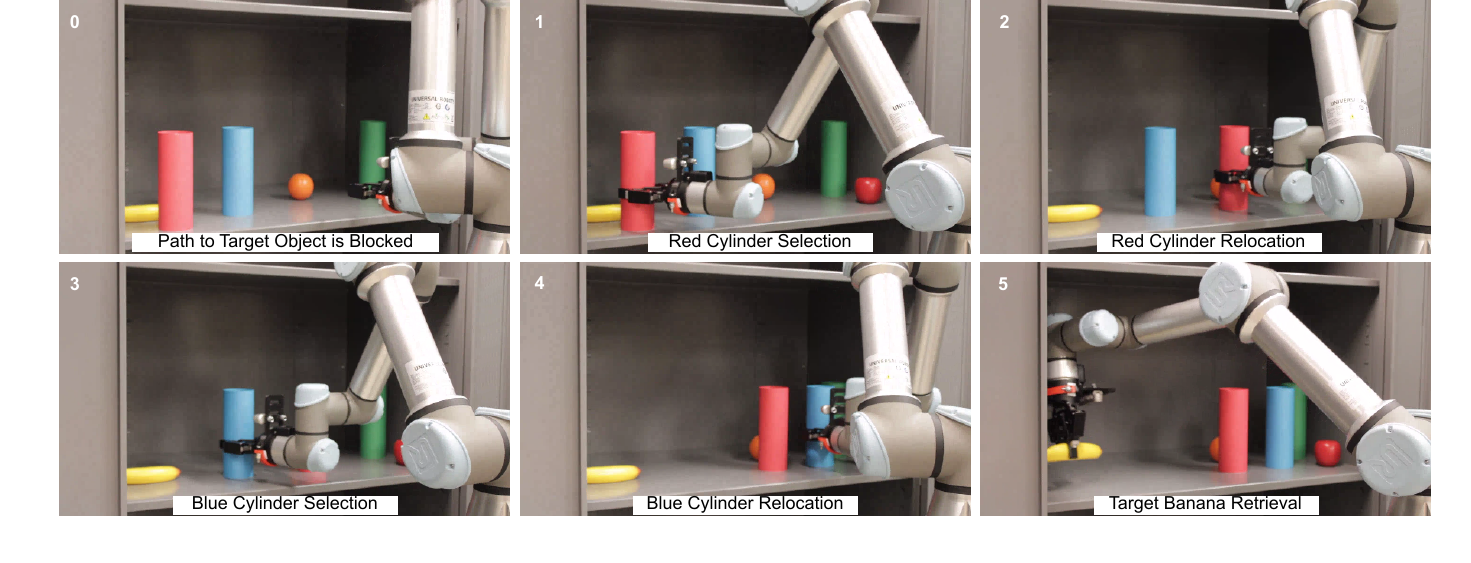}
    \caption{Execution for retrieving the yellow target object (``banana''): In the initial setup, the target object is not retrievable as other objects block it. The robot clears the pathway by moving two cylindrical objects (frames 1-4 ) and then finally take the pathway going through the back of all objects to the target object. It can also be seen that confined spaces impose significant challenges in robot motion, especially when retrieving an object with a relatively lower height, such as a banana, than other objects. }
    \label{fig:real-experiment}
    \vspace{-0.0in}
\end{figure*}
\subsection{Full Pipeline Algorithm}
%Algorithm \ref{alg: neural object retrieval approach} describes our neural rearrangement planning for object retrieval from unknown confined spaces. The robot locates at a fixed location $l_r$. The active neural sensing module is first deployed to gather scene information, from where all objects $o_{\{m+1\}}$ and regions $l_{\{b\}}$ can be extracted from camera perception. Once the target object $o^*$ is successfully detected from $o_{\{m+1\}}$, the robot arm would try to find a path to grasp it using the RRT-connect algorithm. The gripper must get to pose $x_g$ to successfully retrieve the $o^*$ (line 1). If the plan fails, OSNet takes the current scene configuration information consisting of $l_r$, $x_g$, the target object $o^*$, and all other objects $o_{\{m\}}$ and outputs a categorical distribution $\hat{p}_{\{m\}}$ denoting the probability of selecting each object. The object $o'$ with the most significant selection probability will be chosen to be the next one to be rearranged (line 3). In the next step, RPNet kicks in and associates all the regions $l_{\{b\}}$ with a cost value based on $o'$. The best-relocating region $l_{best}$ for $o'$ is obtained by sorting all cost values in ascending order (line 4). Then the robot arm plans a trajectory to move the selection object $o'$ to be desired region $l_{best}$ (lines 5-6).  After object rearrangement, the robot arm will attempt to grasp the target object and repeat the above process (lines 2) until a path solution is found. Finally, the robot grasps the target object and declares success.

Our neural rearrangement planning for object retrieval from unknown confined spaces works as follows. The inputs to our framework include the robot's location $l_r$, a grasp pose $x_g$ for retrieving the target object,  target and non-target objects' states $o_{\{m+1\}}$, and environment surface regions $l_{\{b\}}$. The active sensing module takes the environment dimensions $(d_x, d_y, d_z)$ and constructs the environment to generate the object states $o_{\{m+1\}}$ and surface regions $l_{\{b\}}$. The grasp pose is obtained using GraspNet \cite{sundermeyer2021contact}, and the robot base location $l_r$ is fixed and assumed to be known. Given the inputs, an indicator function leveraging the RRTConnect detects if the target object is reachable by the robot without collision. If not, the algorithm enters the while loop, where it greedily selects and rearranges the non-target objects if such relocation actions are feasible until the pathway toward the target becomes clear, or the loop limit is achieved. A non-target object $o' \in o_{\{m\}}$ is selected for rearrangement using our OSNet output probabilities $p_{\{m\}}$. The $\argmax$ over $o_{\{m\}}$ returns the object with maximum selection probability. Once the object $o'$ is selected, the RPNet proposes the region $l^g_{o'}\in l_{\{b\}}$ for its placement. The region with minimum cost is selected using $\argmin$ over $l_{\{b\}}$. If the path from the selected object's current state $l^s_{o'}$ to propose state $l^g_{o'}$ exists, the robot performs the rearrangement action. Finally, when the path toward the target object is clear, the robot retrieves the object $o^*$ with the given grasp pose; otherwise, our method reports failure when the loop limit is reached.

\begin{table*}[!ht]
  \fontsize{7}{5}\selectfont
  \begin{center}
    \begin{tabular}{c c c c c c}
      \toprule
       & 
      \multirow{2}{*}{\textbf{Object Retrieval Planner}} & \multicolumn{3}{c}{\textbf{Metrics}}\Tstrut\Bstrut \\
       & & Success Rate ($\%$) $\uparrow$ \Tstrut\Bstrut & Objects Rearranged $\downarrow$ & Planning Time (s) $(\%)$ $\downarrow$ & Workspace Moving Distance (m) $\downarrow$\\ \midrule
       & Random Planner \Tstrut\Bstrut & 45$\%$ & 1.689 $\pm$ 0.962 & 0.001 $\pm$ 0.001 & 1.97 $\pm$ 1.204\\
      Comparison & Local Planner \Tstrut\Bstrut & 55 $\%$ & 1.745 $\pm$ 0.976 & 3.307 $\pm$ 1.669 & 1.62 $\pm$ 0.734\\
      %RRT Planner\Tstrut\Bstrut & 71 $\%$ & 1.69 $\pm$ 0.897 & 8.957 $\pm$ 7.735 & 1.575 $\pm$ 0.836\\
     & Neural Planner (Ours)\Tstrut\Bstrut & 72 $\%$ & 1.722 $\pm$ 0.803 & 0.014 $\pm$ 0.006 & 1.647 $\pm$ 0.856\\
      \midrule
      \multirow{3}{*}{Ablation} & OSNet-only Planner \Tstrut\Bstrut & 71$\%$ & 1.662 $\pm$ 0.903 & 2.987 $\pm$ 1.636 & 2.074 $\pm$ 1.185\\
       & RPNet-only Planner \Tstrut\Bstrut & 47 $\%$ & 1.638 $\pm$ 0.783 & 0.01 $\pm$ 0.005 & 1.808 $\pm$ 0.913\\
     %Neural Planner (Ours)\Tstrut\Bstrut & 72 $\%$ & 1.722 $\pm$ 0.803 & 0.014 $\pm$ 0.006 & 1.647 $\pm$ 0.856\\
      \bottomrule
    \end{tabular}
    \caption{The table above shows the statistical comparison between various object rearrangement planners. Overall, our neural planner achieves the highest success rate while keeping the minimum planning times. The bottom table is the result of our ablation study. The higher performance of our neural framework than its ablations validates the effectiveness of our object selection (OSNet) and region proposal network (RPNet) in solving object retrieval tasks.}
    \label{tab:table1}
    \vspace{-0.2in}
  \end{center}
  %\vspace{-0.3in}
\end{table*}

\section{Results \& Discussions}
In this section, we present the results and analysis of the following evaluation experiments: 1) a comparison experiment evaluating the performance of the proposed neural rearrangement planning method against the baselines in unknown confined environments; 2) an ablation study showing the effectiveness of our various neural functions involved in the system; 3) sim-to-real transfer demonstrating our approach's performance in different novel real-world cabinet settings. We use the following metrics for quantitative evaluation.

\begin{itemize}
    \item \textbf{success rate}: It tracks the percentage of cases in the test set where the robot successfully retrieved the target object within the limit of 5 steps. An object selection and relocation is considered as 1 step, so the robot had the limit not to select and relocate more than 5 objects.  
    \item \textbf{object rearranged}: It represents the number of objects rearranged before the target can be successfully retrieved.
    \item \textbf{planning time}: It stores all the time spent selecting the object to be arranged and the time consumed in determining the best region for the selected object's relocation.
    \item \textbf{workspace moving distance}: It shows the total distance the robot arm's end-effector moves to relocate various objects before successfully retrieving the target.
\end{itemize}
\subsection{Baselines}
We create three baseline planners to compare with our method in 100 unseen dynamically generated testing environments with different object numbers, types, and dimensions. All methods start with the scene observation given by the active sensing module comprising object states and surface regions. At the beginning of each step, we run RRTConnect with a fixed time budget of one second as the initial attempt to reach the target object. If it fails, the first baseline, the random planner, randomly selects a non-target object and places it at a randomly selected valid collision-free region. The second baseline is a greedy local planner, greedily relocating all the non-target objects blocking the linear, straight-line path between the robot's current and final states for reaching the target object. The placement region is also chosen greedily based on the nearest collision-free spot next to the selected object, which does not block the linear path trajectory. The baselines and our approach share a maximum step number of 5 before declaring failure. Therefore, the results in Table \ref{tab:table1} are only analyzed for successful experiments for each method based on the metrics mentioned above.

\subsubsection{Success rate}
Overall, the proposed neural method achieves the best success rate among all methods. For the local planner, it tries to remove the objects that block the linear interpolated robot arm path. However, when reaching the target, the arm uses RRTConnect to compute the final trajectory instead of following this linear path because of its often infeasibility in the confined environment setting. As a result, the relocated object may not be optimal for the RRTConnect path planner, which could be one of the reasons for the relatively low success rate. On the other hand, the random planner does not follow any heuristic that can guide it to clear the objects; hence, it mostly fails to retrieve the target objects.
\subsubsection{Planning time}
We introduced neural networks to the object retrieval problem mainly because of their generalization power and fast execution speed at evaluation time. Hence, once trained, our neural model can quickly predict the possible object rearrangement sequences with few forward propagation passes of scene information through our OSNet and RPNet. From Table \ref{tab:table1}, we can see the large planning time gap between our neural method and the local planner. This is because of the potential path calculation and the robot arm trajectory mesh generation for collision checking at every possible placement region. For the random planner, randomly choosing objects and regions is quick, but the success rate is sacrificed.
\subsubsection{number of object rearranged $\&$ moving distance}
Due to the confined environment settings, the difference between the number of objects rearranged and the workspace move distance is not apparent. Our neural planner shared similar values with the baselines. However, since we only show the data analysis for the successful experiments, the small move distance and rearrangement steps for random and greedy planners are mainly caused by many failure cases that proposed relocating more than four objects.

\subsection{Ablation Studies}
In this section, we conduct ablation studies to show the importance and effectiveness of our OSNet and RPNet. All experiments are performed on the same 100 dynamic environments used in the previous section. The results are summarized in Table \ref{tab:table1}. 
\subsubsection{OSNet-only Planner}
In the first study, we replaced our RPNet with the analytical selection module used in the local planner. The results show that the OSNet-only planner still maintains a decent success rate. However, the planning time increases almost a hundred times. Thus, this validates that our RPNet can provide reasonable regions for relocating selected objects and saves significant computational overload when compared to classical methods. 
\subsubsection{RPNet-only Planner}
In the second study, the OSNet is substituted by the classical object selection module from the local planner. According to the results, in the RPNet-only planner, the success rate drops to $47\%$. This evidence shows that our OSNet leads to a significant performance gain over a classical greedy method.  

In summary, our ablation study shows that the ablated models either perform poorly in computational speed or exhibit a lower success rate than our proposed framework, validating the need for both OSNet and RPNet for solving object retrieval tasks. 

\subsection{Real World Experiments}
Finally, we perform a series of real-world experiments in a confined cabinet with dimensions of (56 cm, 86 cm, 50 cm). We directly deploy our neural models trained in the simulated environments to the real robot and object retrieval setup. Two successful trials can be seen in Fig. \ref{fig:real-experiment2} and Fig. \ref{fig:real-experiment}. For the experiment in Fig. \ref{fig:real-experiment}, the target is placed on the left side next to the environment boundary, far away from the initial posing of the robot gripper. We also place two cylinder blocks around the target object to ensure the robot can not directly reach it. The robot starts by performing the active sensing pipeline to understand the initially unknown environment. During the grasping phase, the robot arm first picks the red cylinder directly in front of the target (Fig. \ref{fig:real-experiment}, frame 1) and rearranges it to an adjacent region. In the second step, the robot arm chooses the blue cylinder (Fig. \ref{fig:real-experiment}, frame 3) that blocks the path toward the target and relocates it to a vacant area. At last, after swinging the forearm inside the cabinet, the target object is successfully retrieved by the robot. From the robot's final pose, we can clearly see the reason for choosing the two objects. On the other hand, for the experiment shown in Fig. \ref{fig:real-experiment2}, because of the limited feasible space on the left part of the scene, the planner relocates one of the blocking objects to the left while another to the right near the boundary. Eventually, the target object plum is successfully retrieved.\par
%In the second set of experiments illustrated in Fig. \ref{fig:exp2-figure}, the target is placed on the left side of the scene, with at least one object blocking the gripper from accessing it. This time, the robot goes directly to the blocking cylinder ( Fig. \ref{fig:exp2-figure}, frame 0) and moves it to an empty region not too far away from the initial location. This step clears the space toward the target, and the robot gets into the confined cabinet before successfully retrieving the target object. Unlike the previous experiment's path, the human agent is more likely to agree the final path is optimal for target retrieval. 
%Fig. \ref{fig:exp2-figure} demonstrated another real-world scenario in which the target was placed on the left side with at least one object blocking the gripper from accessing it. In this case, the robot goes directly to the blocking cylinder and moves it to an empty region not too far away from the initial location. This step clears the space toward the target, and the robot gets into the confined cabinet to retrieve the target object successfully. 

Note that the real experiment exhibits the generalizability of our proposed approach to real-world scenarios with direct sim2real transfer. Thus, the successful execution of our approach in these setups validates the effectiveness of our proposed framework.
\section{Conclusions \& Future Work}
This paper presents a neural rearrangement planning method for object retrieval tasks from unknown confined spaces. Our approach can generate a sequence of object selection and their alternative placements in confined spaces to successfully retrieve the given target object. We demonstrate the generalization of the proposed approach to complex real-world scenarios without additional training. Furthermore, the results also show that our framework presents the best performance among all baselines and saves significant computation times in solving object retrieval tasks. Our method achieves such high performance by ensuring the relocation of non-target objects clears the way for the robot path homotopy to the given target object, thus significantly increasing the underlying motion planner's efficiency and chances of finding the required robot motion sequences in confined spaces. %In future work, we would like to further improve the performance of our method by incorporating it with the continual reinforcement learning approaches. Another direction to explore is the tight integration of the active sensing and our object retrieval framework at a step level instead of performing them separately. 

%Currently, despite the best results among all baselines, the overall success rate of our approach is around 72\%, and we believe that it can be enhanced by letting the system learn from its failures in a continual manner. Another direction to explore is the tight integration of the active sensing and our object retrieval framework at a step level instead of performing them separately. The resulting system will enable execution even when the environment is partially observed, leading to a more robust, efficient, and reliable active object retrieval framework.

%However, the main challenge for the hybrid system is that noisy real-world camera perception quality may result in grasp algorithm failures. Therefore, we also plan to explore ways to improve robot grasping under the noisy perception of the given environment.
%%%%%%%%%%%%%%%%%%%%%%%%%%%%%%%%%%%%%%%%%%%%%%%%%%%%%%%%%%%%%%%%%%%%%%%%%%%%%%%%
\bibliographystyle{IEEEtran}
\bibliography{root}

\end{document}